\documentclass{article}
\usepackage{arxiv}

\usepackage[utf8]{inputenc} % Allow utf-8 input
\usepackage[T1]{fontenc}    % Use 8-bit T1 fonts
\usepackage{hyperref}       % Hyperlinks
\usepackage{url}            % Simple URL typesetting
\usepackage{booktabs}       % Professional-quality tables
\usepackage{amsfonts}       % Blackboard math symbols
\usepackage{nicefrac}       % Compact symbols for 1/2, etc.
\usepackage{microtype}      % Microtypography
\usepackage{lipsum}         % Placeholder text
\usepackage{graphicx}       % For including images
\usepackage{natbib}         % Bibliography support
\usepackage{doi}            % DOI support
\usepackage{authblk}        % Author affiliations
\usepackage{amsmath}        % Math symbols and environments
\usepackage{array}          % Advanced table formatting
\usepackage[table,xcdraw,dvipsnames]{xcolor} % Unified xcolor options
\usepackage{soul}           % Text highlighting
\usepackage{geometry}       % Adjust margins
\usepackage{tabularx}       % Full-width tables
\usepackage{multirow}       % Multirow support for tables
\usepackage{longtable}      % Tables spanning multiple pages

\usepackage{makecell}  % For \makecell
\usepackage{tabularx}  % For \tabularx

\title{X-Intelligence 3.0 \\ Training and Evaluating Reasoning LLM for Semiconductor Display}

\date{}

\author{TCL Corporate Research\textsuperscript{*} \and 
	\vspace{-18pt} 
	TCL China Star Optoelectronic Technology Co, Ltd.\textsuperscript{*} \and 
	\vspace{-18pt} 
	National Center of Technology Innovation for Display\textsuperscript{*}}

\renewcommand{\headeright}{Technical Report}
\renewcommand{\undertitle}{}
\renewcommand{\shorttitle}{X-Intelligence 3.0}

\hypersetup{
pdftitle={A template for the arxiv style},
pdfsubject={q-bio.NC, q-bio.QM},
pdfauthor={David S.~Hippocampus, Elias D.~Striatum},
pdfkeywords={First keyword, Second keyword, More},
}

\begin{document}
\maketitle
\vspace{-10pt}

\renewcommand{\thefootnote}{*}
\footnotetext{Authors who contributed to this work can be found on page~\pageref{Contributors_Acknowledgments}.
}

\begin{abstract}
	
Large language models (LLMs) have recently achieved significant advances in reasoning and demonstrated their advantages in solving challenging problems. Yet, their effectiveness in the semiconductor display industry remains limited due to a lack of domain-specific training and expertise. 
To bridge this gap, we present X-Intelligence 3.0, the first high-performance reasoning model specifically developed for the semiconductor display industry.
This model is designed to deliver expert-level understanding and reasoning for the industry's complex challenges.
Leveraging a carefully curated industry knowledge base, the model undergoes supervised fine-tuning and reinforcement learning to enhance its reasoning and comprehension capabilities.
To further accelerate development, we implemented an automated evaluation framework that simulates expert-level assessments.
We also integrated a domain-specific retrieval-augmented generation (RAG) mechanism, resulting in notable performance gains on benchmark datasets.
Despite its relatively compact size of 32 billion parameters, X-Intelligence 3.0 outperforms SOTA DeepSeek-R1-671B across multiple evaluations. This demonstrates its exceptional efficiency and establishes it as a powerful solution to the longstanding reasoning challenges faced by the semiconductor display industry.

\end{abstract}

% \will{The abstract should show the contribution in a high-level way}

% keywords can be removed
\keywords{Semiconductor Display \and Domain-Specific Reasoning LLM \and Domain-Specific Post-Training \and Domain-Specific Retrieval-Augmented Generation \and Language Model-Based Automated Evaluation Method}

\section{Introduction}

\begin{figure}[h!]
	\centering
	\includegraphics[width=0.95\textwidth]{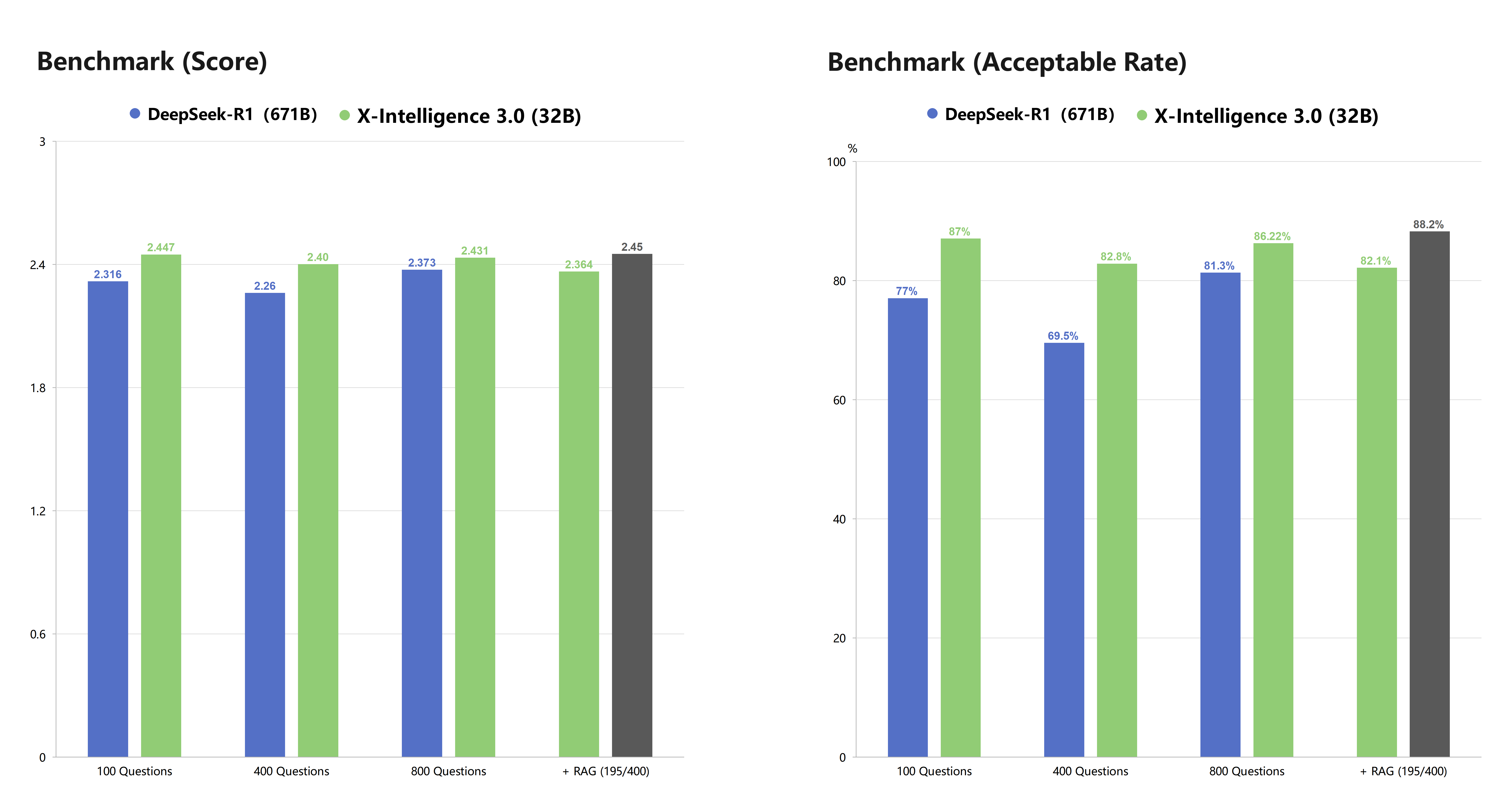} 
	\caption{Benchmark performance of X-Intelligence 3.0.}
	\label{fig:F_1}
\end{figure}

Semiconductor display technology combines semiconductor processes with optoelectronic devices, using thin-film transistors (TFTs) to drive display units (emissive or modulative) for information visualization. This technology has evolved from cathode ray tubes (CRT) \cite{menozzi2001crt} to plasma displays (PDP) \cite{sobel1991plasma}, liquid crystal displays (LCD) \cite{schadt1997liquid}, organic light-emitting diodes (OLED) \cite{song2020organic}, quantum dot OLEDs (QD-OLED) \cite{dayneko2016application}, Mini- and Micro-LEDs \cite{huang2019prospects}, and most recently to silicon-based Micro-OLED and Micro-LCD microdisplay technologies \cite{miao2024microdisplays}.
This evolution consistently targets key goals: higher resolution, lower power consumption, better display quality, and greater flexibility in form factors. The industry relies on close collaboration across the value chain. Upstream focuses on core materials and equipment—such as organic emitters, quantum dots, liquid crystals, and vapor deposition tools—to enhance performance. Midstream leverages precision processes like TFT array fabrication, color filters, alignment, and encapsulation to achieve submicron-level manufacturing. Downstream, driver ICs and module designs support high-brightness, ultra-high contrast, and flexible displays for smartphones, automotive, and AR/VR applications.
As display devices continue to shrink and complexity rises, the demand for advanced manufacturing and intelligent design grows rapidly. In this context, artificial intelligence (AI) is being deeply integrated into the display development pipeline, optimizing process parameters, predicting defects, and assisting in circuit and pixel design, accelerating technological progress in the display industry.

Recent breakthroughs in the reasoning capabilities of large language models (LLMs) have become a central focus in AI research \cite{plaat2024reasoning, xu2025towards}. 
OpenAI’s o1 model \cite{openai2024a}, released in 2024, significantly improved the ability to solve complex tasks through an "exploration–reflection–iteration" reasoning process.
DeepSeek-R1 \cite{guo2025deepseek} leveraged a verifiable reinforcement learning strategy, training on verifiable datasets such as mathematics and code with minimal cold-start samples and a staged training framework. This approach not only demonstrated emerging reasoning abilities but also markedly increased answer accuracy, emphasizing the value of reinforcement learning in enhancing large-scale model reasoning.
Anthropic released the first hybrid reasoning model, Claude 3.7 Sonnet \cite{anthropic2025claude37}, which can switch between rapid response and deep thinking modes.
Similarly, Alibaba's open-source Qwen 3 series \cite{yang2025qwen3} incorporated a hybrid reasoning mechanism, enabling manual toggling between reasoning and non-reasoning states to optimize computational efficiency.
Meanwhile, Google’s Gemini 2.5 Pro \cite{google2025gemini25}, through enhanced post-training optimization, topped the Chatbot Arena leaderboard \cite{chatbot_arena_leaderboard}, further pushing the frontier of model performance.

However, directly applying general reasoning models to the semiconductor display industry faces significant challenges due to the industry's highly vertical and process-intensive nature \cite{verma2025framework}. Reasoning tasks in this field require the integration of multidisciplinary knowledge across materials, processes, equipment, devices, circuit design, and IC drivers, along with optical and electrical inspection data for further analysis and decision-making. 
This not only demands deep interdisciplinary information fusion but also depends on step-by-step decision-making logic. 
In practice, deploying large language models in the semiconductor display industry often encounters the following core issues:

\textbf{Fragmented knowledge:} The industry's complexity and the dispersion of information across numerous sources hinder the formation of a unified knowledge system, reducing integration and application efficiency.

\textbf{Difficult data annotation:} Labeling data requires deep domain expertise, making the process costly and time-consuming and resulting in a shortage of high-quality, labeled datasets.

\textbf{Data security risks:} Core enterprise data involves trade secrets, requiring careful handling of both general and proprietary information.

\textbf{Lengthy iteration and validation:} The lack of domain-specific evaluation methods leads to heavy reliance on manual expert reviews, which are both costly and inefficient.

\textbf{Poor generalization:} With more than 820 subdomains in our four-level taxonomy, models often underperform when encountering unseen knowledge points.

To address these challenges, we present X-Intelligence 3.0, the first high-performance reasoning model specifically developed for the semiconductor display industry. This model is designed to overcome the limitations of general LLMs by constructing a comprehensive domain-specific dataset and employing a targeted training methodology that integrates deep industry knowledge. Our main contributions include:
\begin{itemize}
	\item \textbf{Semiconductor Display Dataset:} We developed a high-quality data-generation pipeline for post-training. This dataset combines open-source data, expert annotations, curated data collected via the X-Intelligence platform, retrieval-augmented generation (RAG) synthesis, and distilled content from technical literature and textbooks. All data is rigorously filtered to ensure strong relevance and accuracy for the semiconductor display domain.
	\item \textbf{First Reasoning Model for Semiconductor Display:} X-Intelligence 3.0 is the first reasoning model tailored specifically to the semiconductor-display industry. It achieves state-of-the-art results across multiple domain-specific benchmarks, outperforming the DeepSeek-R1-671B model.
	\item \textbf{Customized Domain RAG Methodology:} We introduce a specialized RAG mechanism that reliably supplies domain-relevant knowledge while reducing hallucinations, yielding significant performance gains on the benchmarked dataset.
	\item \textbf{Automated Evaluation Framework:} To support rapid iteration and reliable assessment, we built an LLM-driven evaluation platform that emulates expert review. This system employs proprietary evaluation sets containing 100 and 400 domain-specific questions, enabling comprehensive and scalable model validation.
\end{itemize}

\section{X-Intelligence: Industry-Specific LLM for Semiconductor Display}
\label{data_curation_section}

Developing large language models customized to specific domains requires deep expertise in the target field.
The mainstream approach follows a comprehensive "pre-training + post-training" pipeline. 
X-Intelligence 1.0 \cite{tcl2023x-intelligence1.0} successfully implemented this framework, surpassing GPT-4 \cite{GPT4} and demonstrating its effectiveness in the semiconductor display industry.
To further improve performance and enhance reasoning capabilities, this paper introduces X-Intelligence 3.0, which places greater emphasis on the post-training phase.
Specifically, we explore how to fine-tune a general reasoning model with high-quality, domain-specific data and, through a carefully devised reinforcement-learning strategy, cost-effectively unlock the model’s inherent reasoning potential, resulting in significant improvements.
As shown in Fig. \ref{fig:F_1}, X-Intelligence 3.0 delivers state-of-the-art results, outperforming DeepSeek-R1-671B  \cite{guo2025deepseek} across multiple evaluations. 
The following section provides a detailed overview of the X-Intelligence 3.0 training pipeline.

\subsection{Supervised Fine-tuning}
The X-Intelligence-3.0 is fine-tuned using both domain-specific and general-domain data, augmented with Chain-of-Thought (CoT) \cite{wei2022chain} reasoning. Domain-specific data is primarily constructed through an automated pipeline, with limited involvement from human experts. This process is divided into two stages: Stage 1 focuses on question generation, while Stage 2 involves CoT generation and answer distillation using DeepSeek-R1. General-domain data is sourced from publicly available datasets and constructed through thorough screening and refinement.

\textbf{Domain-Specific Question Construction.} Domain-specific questions are sourced from three distinct origins:
(1) User data accumulated within internal systems, representing real-world production issues.
(2) High-quality questions written by experts across several semiconductor display sub-domains.
(3) Questions extracted by LLMs from internal high-value, business-critical technical documents and external cutting-edge research papers.
After construction, all questions undergo deduplication and filtering. For deduplication, the BGE-M3 model \cite{chen2024bge} computes semantic vectors for each query. Query pairs with cosine similarity exceeding 0.9 are considered duplicates, and only one instance is retained. During filtering, LLMs eliminate incomplete, ambiguous, or non-semiconductor-display-related questions.

Extensive literature research indicates that instruction fine-tuning data with excessively low complexity yields negligible performance gains. 
In contrast, high-complexity data significantly enhances model generalization on challenging tasks. 
Consequently, we perform complexity enhancement on questions to improve the model's ability to handle complex problems. Specifically, we first use an LLM to identify high-value but low-complexity questions from the constructed set. 
This subset is then rewritten and enhanced by an LLM to produce more challenging versions. 
This process undergoes iterative refinement through data sampling and expert validation until the desired level of complexity is achieved.

\textbf{Domain-Specific Answer Distillation.} After question generation, for questions with standard answers or source slices, we select Deepseek-R1 \cite{guo2025deepseek} as the teacher model for data distillation. For questions without standard answers or source slices, we adopted a multi-model generation and multi-strategy fusion scheme to ensure the accuracy of the distilled thought chains and answers.

Deepseek-R1 \cite{guo2025deepseek} and OpenAI-o1 \cite{openai2024a} are selected as teacher models for multiple data distillation, that is, each query was used to automatically generate multiple thought chains and answer pairs.
Deepseek-V3 \cite{liu2024deepseek} was used to automatically evaluate the generated thought chain–answer pairs, and those with the highest evaluation scores were selected as the final outputs.
To ensure the reliability of automated evaluation, experts were assigned to conduct manual synchronous evaluations during the initial stage.
Based on the results of expert evaluations, we optimized Deepseek-V3’s automated evaluation strategy. A high-quality domain-specific supervised fine-tuning (SFT) dataset featuring chain-of-thought annotations was ultimately produced, serving as foundational support for subsequent instruction tuning training.

\textbf{General-Domain Data Screening.} To preserve the general capabilities of our models during fine-tuning, we designed a method to screen high-quality general-domain data, considering its diversity, validity, accuracy, and complexity.

Diversity: Data was collected from multiple open-source corpora \cite{huggingface_open_r1, tian2025deepdistill, penedo2025codeforcescots} covering tasks including mathematical reasoning, code generation, scientific reasoning, instruction following, and others.

Validity: Raw queries were filtered to remove duplicates, incomplete entries, hyperlinks, and contaminated data.

Accuracy: Deepseek-R1 was used to distill answers from collected data, and model-generated answers are verified against ground truth. 
Data with verification scores falling below a predefined threshold was discarded.
The verification scores of different categories are calculated as follows:

(1) Math: Math Verify was used to evaluate the model’s outputs and assign binary correctness labels.

(2) Code: The verification score was computed based on sandboxed execution results across selected test cases.

(3) Science and others: Qwen3-32B was used to evaluate the similarity between generated results and ground truth answers.

(4) Instruction following: Instruction-following (IF) data was evaluated using the ifeval validator. The verification score was computed as follows:

\begin{equation}
	\text{verify\_score}_{\text{IF}} = \frac{1}{m} \sum_{i=1}^{m} \text{pass\_score\_constraint}_i, \quad \text{where } \text{pass\_score\_constraint}_i \in \{0, 1\}
\end{equation}

(5) Challenge: compute perplexity (PPL) and difficulty score (on a 1-5 scale) using Deepseek-R1, and calculate CQD (Complexity Quality Difficulty) using Eq. \ref{CQD}. Based on the resulting CQD score, data was categorized into advanced $(\text{CQD} \geq 0.8)$, intermediate $(0.5 \leq \text{CQD} < 0.8)$, and simple $(\text{CQD} < 0.5)$ levels. Data was selected from the advanced and intermediate categories as required.

\begin{equation}\label{CQD}
	\text{CQD} = \alpha \left(1 - \frac{\text{PPL} - \text{PPL}_{min}}{\text{PPL}_{max} - \text{PPL}_{min}}\right) + \beta \frac{\text{difficulty\_score} - 1}{5 - 1}
\end{equation}

\subsection{Reinforcement Learning}

\begin{figure}[h!]
	\centering
	\includegraphics[width=0.95\textwidth]{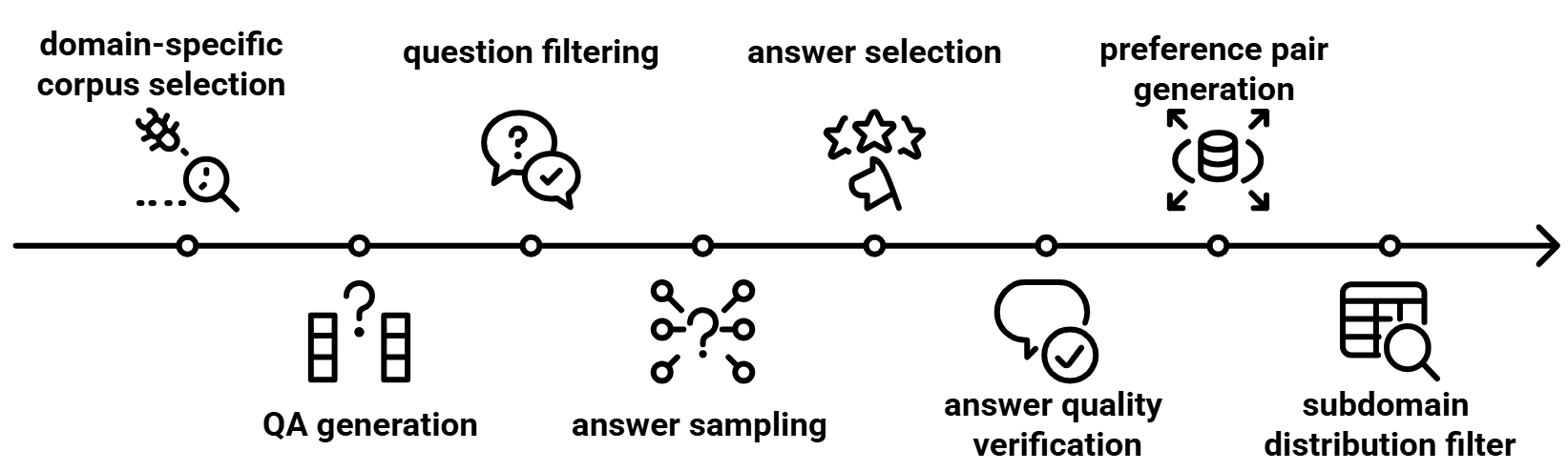} 
	\caption{Pipeline for vertical-domain preference data generation.}
	\label{fig:dpo}
\end{figure}

To further enhance the model's reasoning capabilities, we performed reinforcement learning fine-tuning using vertical-domain preference data pairs. Due to the lack of readily available open-source datasets for semiconductor display, we developed an automated vertical-domain preference data generation pipeline and used it to train the model as shown in Fig. \ref{fig:dpo}.

\textbf{QA Generation. }
We utilized academic papers and dissertations in the field of semiconductor display as the primary source for reinforcement learning data, as these materials exhibit strong domain-specific expertise, accuracy, and logical reasoning potential. Firstly, we used Qwen3-32b model \cite{yang2025qwen3} to extract challenging reasoning questions based on the paper content. These questions were required to meet the following criteria: (1) precise description of the problem, (2) existence of a definitive answer, and (3) derivability of the answer through logical reasoning from the source paper. The model was then tasked with generating reference answers by strictly adhering to the original paper content.

Since different papers might yield semantically similar questions, we employed the simHash algorithm \cite{charikar2002similarity} to perform word-level similarity calculations across all generated questions. A threshold-based filtering approach was applied: questions with similarity scores below 0.7 were retained directly, those above 0.9 were discarded, and questions falling between 0.7 and 0.9 underwent further semantic-level deduplication using the large language model.

To further enhance question quality, we leveraged the Qwen3-32b model to filter out : (1) queries that are not easily verifiable, (2) simple problems requiring no substantive reasoning, or (3) queries that lacked generalizability or universal applicability.

\textbf{Preference Data Generation. }
After finalizing the question set, we employed a high sampling temperature to prompt the SFT model to generate multiple diverse responses for each question. The Qwen3-32b model was then used to evaluate these responses against the reference answer, selecting the best and worst responses for each question. To ensure the reliability of this relative quality assessment, an additional iterative pairwise comparison with randomized ordering to eliminate positional bias is performed to confirm the superiority of the chosen response over the rejected one.

To mitigate the impact of absolute answer quality, we further filtered the data based on the scores of chosen responses. Specifically, the Qwen3-32b model scored each chosen response on a 10-point scale relative to the reference answer, and responses scoring below 6 were discarded.

Finally, to ensure balanced data distribution across subdomains, we applied domain-specific labeling to the retained questions. The Qwen3-32b model assigned predefined category labels (prepared through manual annotation) to each question, and the dataset was filtered accordingly to maintain equitable representation of all subfields.

\textbf{DPO Training. }
The curated preference data pairs are then used to perform DPO (Direct Preference Optimization) \cite{rafailov2023direct} training on the SFT model. The DPO optimization objective is:

\begin{equation}
	L_{\text{DPO}}(\pi_\theta; \pi_{\text{ref}}) = -\mathbb{E}_{(x, y_w, y_l) \sim D} \left[ \log \sigma \left( \beta \log \frac{\pi_\theta(y_w \mid x)}{\pi_{\text{ref}}(y_w \mid x)} - \beta \log \frac{\pi_\theta(y_l \mid x)}{\pi_{\text{ref}}(y_l \mid x)} \right) \right]
\end{equation}

\section{Retrieval-augmented generation}

Vertical domain knowledge exhibits characteristics of high specialization, terminological density, and complex logical relationships, causing generic RAG frameworks \cite{huang2024survey} to suffer from insufficient retrieval precision and disconnected multi-hop reasoning. This chapter proposes three innovative solutions to significantly enhance domain-specific knowledge retrieval efficacy.

\subsection{Domain-Adaptive Embedding and Rerank Model Fine-tuning}

In vertical domains, generic embedding and rerank models often fail to accurately capture domain-specific semantic features and relevance relationships. For instance, in the semiconductor display domain, professional terminology and issue-resolution logic substantially differ from general domains. Direct application of generic models introduces semantic gaps that degrade retrieval accuracy and impede access to precise domain knowledge. Fine-tuning these models enables better alignment with domain-specific data characteristics, thereby improving retrieval and ranking performance.

\textbf{Automated Construction of Hard Negative Samples. }
Hard negatives refer to samples semantically similar to positives but functionally distinct, a prevalent phenomenon in vertical domains. For example, in legal domains, different statutes may address analogous scenarios but yield divergent outcomes. Weighted training on such samples enhances model discrimination capability for ambiguous cases.

To address low discriminability between positives and negatives in vertical domains, we establish a multi-strategy negative generation framework.
%as illustrated in Fig. \ref{fig:rag_1}.

(1) BM25 Hard Negatives: Extract segments from non-adjacent passages within the same document with >70\% lexical overlap but semantic irrelevance (e.g., "Micro-OLED" vs. "Micro-LED").

(2) Cross-Domain Semantic Negatives: Retrieve semantically similar sentences across sub-domains using semantic models (e.g., retrieving LCD-related negatives from OLED corpora)

(3) Adversarial Negatives: Generate perturbed text through domain-specific LLM paraphrasing.

%\textbf{Dynamic Weighted Training Strategy. }
The loss function incorporates two primary components:

(1) Contrast Loss: A standard contrastive learning objective that minimizes the distance between semantically similar sentences while maximizing separation for dissimilar pairs.

(2) CoSENT Loss: An additional optimization target that further refines model performance.

\textbf{Periodic Hard Negative Mining \cite{li2024conan}: }
At every 1,000 training steps, hard negatives are identified through loss analysis. These challenging samples are subsequently integrated into the training dataset. Repeated exposure to such samples enhances the model's ability to discriminate between low-similarity sentences, thereby improving both discriminative capability and generalization performance.

\subsection{Iterative Retrieval Framework}
\begin{figure}[h!]
	\centering
	\includegraphics[width=0.95\textwidth]{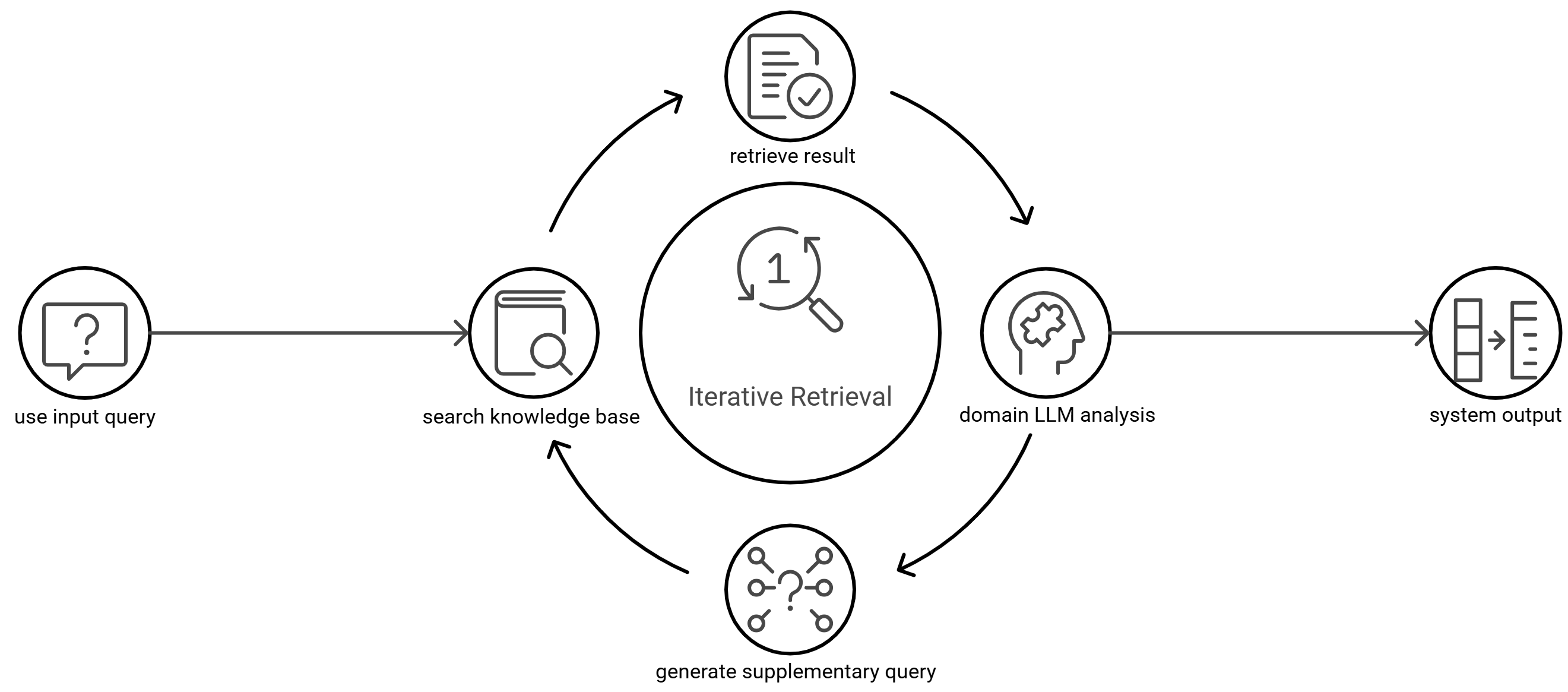} 
	\caption{Iterative Retrieval Framework.}
	\label{fig:rag_3}
\end{figure}

To address the low coverage of single-retrieval for complex queries, we propose an iterative retrieval workflow that incrementally enhances the quality of retrieved content. 
The overall workflow is illustrated in Fig. \ref{fig:rag_3}.
For complex user questions, it is often impossible to obtain complete answer information through a single search. Iterative search gradually improves the search content through the following steps:

\textbf{Initial Retrieval:} The user query is processed through domain-optimized embedding and rerank models to obtain relevant documents.

\textbf{Content Analysis and Expansion:} Retrieved content is analyzed by our domain-adapted 32B LLM, which identifies information gaps and generates supplementary keywords/queries.

\textbf{Secondary Retrieval:} Newly generated queries initiate subsequent retrievals.

\textbf{Iterative Refinement:} Steps 2-3 repeat until comprehensive content coverage is achieved.

\subsection{Automated RAG-SFT Data Construction}
\begin{figure}[h!]
	\centering
	\includegraphics[width=0.95\textwidth]{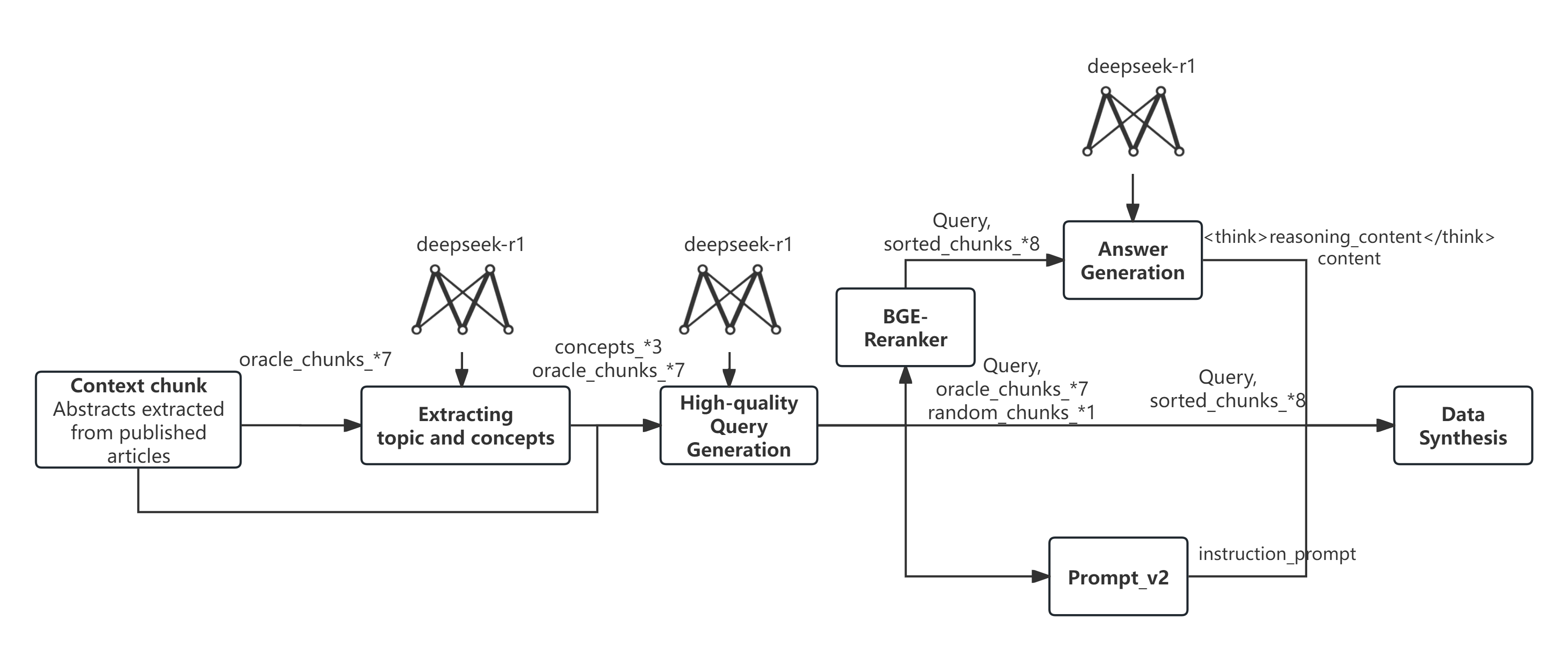} 
	\caption{Automated RAG-SFT Data Construction.}
	\label{fig:rag_4}
\end{figure}

This paper proposes an automated pipeline for constructing RAG-SFT data \cite{qin2025scaling} to extract key information from academic papers and generate high-quality answers. As illustrated in Fig. \ref{fig:rag_4}, the process comprises six stages:

\textbf{Context Chunk Abstraction:}
Original context chunks are extracted from articles as foundational information units.

\textbf{Topic and Concept Extraction:}
A large model (DeepSeek-R1) processes these chunks to identify core topics and concepts. Each chunk contains 5 high-quality oracle chunks and 5 model-generated topic concepts, providing semantic foundations for query generation.

\textbf{Query Generation:}
Leveraging the same DeepSeek-R1 model, this stage synthesizes extracted concepts and oracle chunks to generate precise queries that target essential information for retrieval.

\textbf{BGE-Reranker with Chunk Mixing:}
Queries are processed by a vertical domain-optimized BGE-Reranker. Before reranking, oracle chunks (×5) and random chunks (×3) are intentionally blended to enhance data diversity and simulate real-world noise. The output is reranked chunks (×8) paired with the query.

\textbf{Answer Generation:}
The DeepSeek-R1 model synthesizes the query and reranked chunks using instructional prompts to produce logically coherent answers.

\textbf{Data Synthesis:}
Queries, sorted chunks, and tuned prompts are integrated into a comprehensive dataset for system training and evaluation.

\section{Proposed Evaluation Method for Semiconductor Display}
As our domain-specific large language model is primarily designed to support research, production, and quality inspection within the semiconductor display industry, conventional evaluation frameworks for large models are insufficient due to the absence of specialized benchmarks for this field. To address this limitation, we developed a dedicated evaluation dataset in parallel with the model’s training corpus, combining data collection, expert curation, and automated generation techniques.

The evaluation dataset includes both objective items, such as true/false, multiple-choice, and fill-in-the-blank questions, and subjective, open-ended questions. 
Objective items are well-suited for automated scoring and are effective in assessing factual accuracy and structural compliance, but they offer only a narrow view of the model’s overall capabilities.
In contrast, open-ended questions enable a more comprehensive evaluation of key language competencies, including logical coherence and reasoning. Therefore, our evaluation framework prioritizes open-ended QA tasks for assessing domain-specific proficiency.

\subsection{Human Evaluation}
\label{Human_Evaluation}
Expert human evaluation is essential for accurately assessing the quality of subjective responses generated by the model. 
To ensure consistency and structure in this process, we established six evaluation criteria, each targeting a distinct aspect of model performance:

\textbf{Grammatical Fluency:}
Assesses the mode's fundamental language abilities, including sentence completeness, accurate word usage, and overall fluency.

\textbf{Safety:}
Evaluates whether the response adheres to legal, ethical, and content safety standards, ensuring it is free from harmful or inappropriate material.

\textbf{Logical Reasoning: }
Measures the coherence and soundness of the model's reasoning process, reflecting its capacity for logical thought.

\textbf{Accuracy: }
Determines the correctness of the response, focusing on factual precision and avoidance of hallucinated or fabricated information.

\textbf{Comprehensiveness: }
Examines the model's ability to elaborate on relevant concepts and provide clear explanations of key terms.

\textbf{Practicality: }
Judges how well the response addresses the user's query and whether it offers actionable, relevant, or useful insights.

All criteria, except for Safety, are rated on a 0-3 scale by domain experts, with higher scores indicating better performance. Safety is evaluated using a binary scheme: a score of 3 indicates safe content, while any unsafe output results in a score of 0.

To reflect the varying importance of each dimension in real-world applications, we apply the weights illustrated in Table \ref{tab:weight}, which are informed by expert input and consistent with established evaluation practices. 
The final score for each response is calculated as a weighted sum across all criteria.

\begin{table*}[h!]
	\centering
	\begin{tabularx}{\textwidth}
		{|>{\centering\arraybackslash}X
			|>{\centering\arraybackslash}X
			|>{\centering\arraybackslash}X
			|>{\centering\arraybackslash}X
			|>{\centering\arraybackslash}X
			|>{\centering\arraybackslash}X|}
		\hline
		Grammatical Fluency & Safety & Logical Reasoning & Accuracy& Comprehensive -ness& Practicality \\
		\hline
		10\% & 10\% & 10\% & 20\% & 20\% & 30\% \\
		\hline
	\end{tabularx}
	\caption{Weight of the evaluation criteria.}
	\label{tab:weight}
\end{table*}

In addition to evaluating performance on individual capability dimensions, users also pay close attention to the overall balance of a model’s abilities, preferring systems that do not exhibit marked weaknesses in any particular area.
To address this concern, we worked with domain experts to introduce a new metric—the "Acceptable Rate"—that provides a more holistic view of model performance. 
This metric is defined as the proportion of responses whose accuracy, comprehensiveness, and usefulness scores are all at least 2, indicating that the answer satisfies the minimum quality threshold across all three criteria.

\begin{equation}
	\text{Acceptable Rate} = \frac{N_{\text{acc} \geq 2,\ \text{comp} \geq 2,\ \text{prac} \geq 2}}{N_{\text{total}}}
\end{equation}

This human evaluation framework serves as the standard for all subsequent experiments. 
All reported findings and performance assessments are based on these expert-assigned scores.

\subsection{Objective Evaluation}
\label{Objective_Evaluation}

\begin{figure}[h!]
	\centering
	\includegraphics[width=0.95\textwidth]{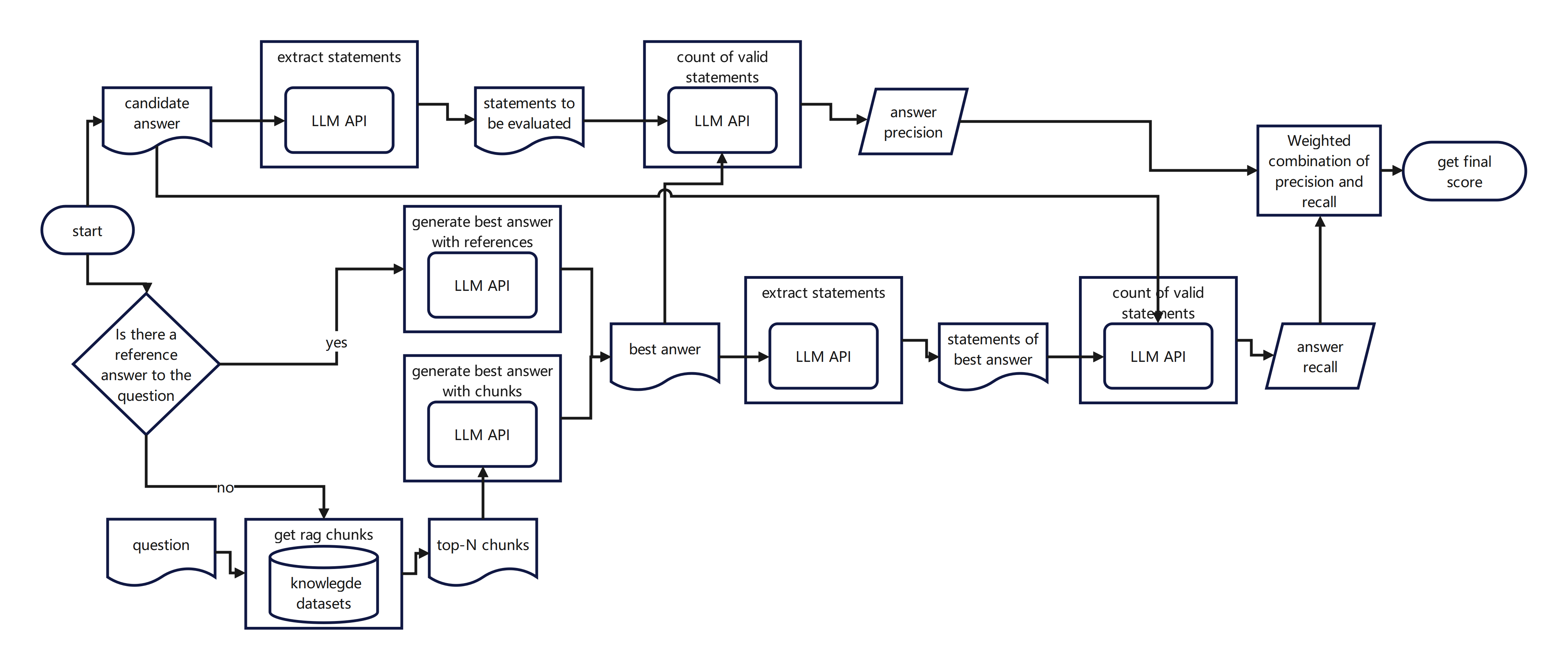} 
	\caption{Workflow of the automatic evaluation framework.}
	\label{fig:subject_eval}
\end{figure}

Subjective evaluations are well-suited for assessing the domain-specific capabilities of language models.  
However, they come with inherent limitations such as high labor costs, low efficiency, and variability in human judgment. 
Evaluators may interpret scoring criteria differently, leading to subjective bias and inconsistency in results.
To overcome these limitations, we propose an automatic evaluation framework based on large language models, aimed at enhancing both the efficiency and reliability of model assessments. This approach is particularly well-suited for iterative model development and large-scale evaluation of open-ended responses.

Our approach is inspired by automatic evaluation methods like Ragas \cite{es2025ragasautomatedevaluationretrieval}. Assuming access to a correct reference answer (Ground Truth), we first decompose the model's response into discrete, concise information units, referred to as statements in our framework. Using LLMs such as OpenAI-o4-mini \cite{openai2025o4mini} and DeepSeek-R1 \cite{guo2025deepseek}, we then determine whether each key point can be logically inferred from the reference answer. The proportion of inferred points serves as our primary metric, \textbf{Answer Precision}, which conceptually aligns with the notion of faithfulness in the Ragas methodology. It is calculated as:

\begin{equation}
	\text{Answer Precision} = \frac{N_{\text{correct-in-resp}}}{N_{\text{resp}}}
	\label{eq:answer_precision}
\end{equation}

$N_{\text{correct-in-resp}}$ means the count of correct statements extracted from model's response, $ N_{\text{resp}} $ means the total number of statements.
	
A higher precision score indicates stronger factual alignment between the model's response and the ground truth, whereas hallucinated or irrelevant content naturally lowers the score, making this metric a direct measure of response accuracy.

However, answer precision alone cannot capture the full utility of a response. For instance, a model might produce factually accurate but contextually irrelevant answers, resulting in high precision but limited usefulness. To address this, we introduce a complementary metric: \textbf{Answer Recall}.

In this reverse process, we decompose the reference answer into statements and evaluate how many of them can be inferred from the model's response. The resulting recall score reflects the extent to which the model covers the essential content of the reference answer:

\begin{equation}
	\text{Answer Recall} = \frac{N_{\text{recalled-from-gt}}}{N_{\text{gt}}}
	\label{eq:answer_recall}
\end{equation}

$N_{\text{recalled-from-gt}}$ means the count of recalled statements extracted from ground truth, and $N_{\text{gt}}$is the total number of statements.

High recall indicates that the response adequately addresses the core elements of the query and is therefore more likely to be useful and complete. This aligns with the comprehensiveness dimension commonly evaluated in human assessments.

Reference answers often vary in quality and style, as they may originate from expert-written content or be synthesized from multiple sources. To ensure consistency, we use large language models to standardize, refine, and consolidate these responses. Where applicable, expert reviewers validate the processed answers to guarantee clarity and reliability. The finalized references are then used as ground truth in the automated evaluation pipeline.

To generate an overall evaluation score, we combine \textbf{Answer Precision} (P) and \textbf{Answer Recall} (R) using a weighted scheme, inspired by the structure of human evaluations:

\begin{equation}
	\text{Final Score} = \alpha \cdot P + \beta \cdot R
	\label{eq:final_score}
\end{equation}

Here, the weights are set to $\alpha = 0.3$ and $\beta = 0.7$, reflecting the greater emphasis on completeness and practical value in real-world use cases, based on expert feedback and existing evaluation practices.
%\begin{flushleft}
	
%\end{flushleft}
With this, a relatively comprehensive automatic evaluation system can operate without human intervention. Since the entire process can be executed programmatically, we can further improve evaluation efficiency through parallel computing or multi-machine deployment. The overall workflow is illustrated in Fig. \ref{fig:subject_eval}.

During validation, we observed that LLM outputs may vary across runs due to inherent randomness, resulting in slight inconsistencies in statement extraction and inference judgments. This variability resembles the subjective noise often seen in human evaluations. To mitigate this, we implemented two stabilization strategies: (1) statements are extracted once and reused across evaluations, and (2) the LLM's temperature parameter is fixed at 0 or 0.1 during inference to reduce randomness.

These measures substantially improve the reproducibility and stability of evaluation results, enabling dependable deployment of the automatic assessment system at scale.

\section{Experimental Result}
\label{experiment_section}

\subsection{Implementation details}
We selected Qwen3-32B \cite{yang2025qwen3} and R1-32B \cite{guo2025deepseek} as the base models for training due to their strong foundational capabilities. To train X-Intelligence 3.0, we utilized 16 NVIDIA A100 GPUs, each with 80 GB of memory.

The hyperparameters for supervised fine-tuning (SFT) include a global batch size of 64, a learning rate of 1e-5, and training for 4 epochs. A cosine learning rate scheduler is used, with a warm-up ratio of 5\%, after which the learning rate decays to 0.
For reinforcement learning (RL), the hyperparameters include a global batch size of 64, a learning rate of 1.0e-4, and training for 1 epoch. A cosine learning rate scheduler is also employed, with 30 warm-up steps, followed by a decay of the learning rate to 1e-6.

\begin{figure}[h!]
	\centering
	\includegraphics[width=0.95\textwidth]{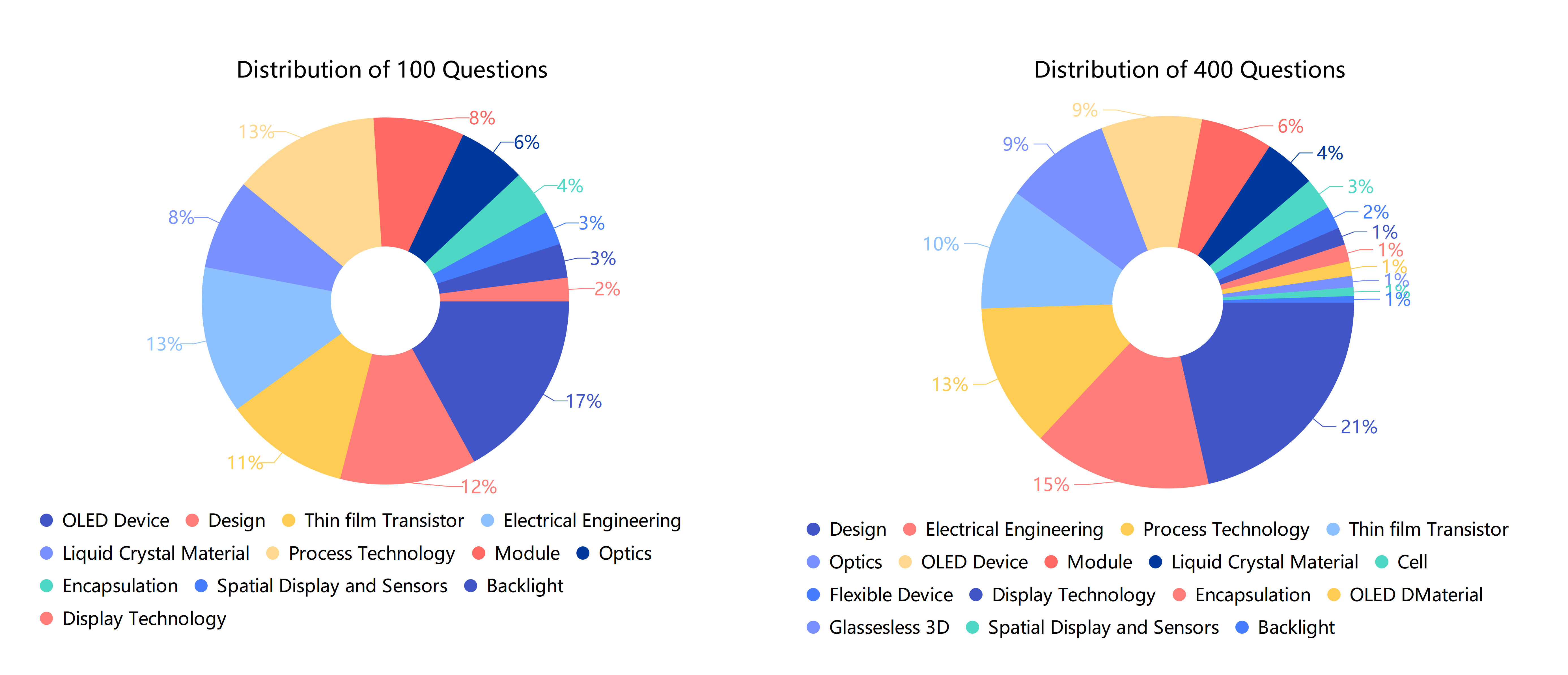} 
	\caption{Distribution of benchmark datasets.}
	\label{fig:F_2}
\end{figure}

To comprehensively evaluate the quality of the model's outputs, we designed three evaluation sets containing 100, 400, and 800 questions, respectively. The 100-question and 400-question sets are non-overlapping, while the 800-question set includes all 400 questions from the latter. The detailed distribution of test questions is shown in Fig. \ref{fig:F_2}.

\subsection{Human Evaluation}

\begin{table*}[h!]
	\centering
	\begin{tabularx}{\textwidth}{l *{3}{>{\centering\arraybackslash}X}}
		\toprule
		\textbf{100 Questions} & \textbf{Deepseek-R1-671B} & \textbf{\makecell{X-Intelligence 3.0 \\ (Trained on R1-32B)}} & \textbf{\makecell{X-Intelligence 3.0 \\ (Trained on Qwen3-32B)}} \\ \midrule
		
		Score & 2.316 & 2.44 & \textcolor{red}{2.447} \\
		
		Acceptable Rate & 77\% & 82\% & \textcolor{red}{87\%} \\
		
		\bottomrule
	\end{tabularx}
	\caption{The performance of our proposed method on the 100 questions benchmark.}
	\label{tab:human_100}
\end{table*}

\begin{table*}[h!]
	\centering
	\begin{tabularx}{\textwidth}{l *{2}{>{\centering\arraybackslash}X}}
		\toprule
		\textbf{400 Questions} & \textbf{Deepseek-R1-671B}& \textbf{X-Intelligence 3.0} \\ \midrule
		
		Score & 2.26  & \textcolor{red}{2.4} \\
		
		Acceptable Rate & 69.5\%  & \textcolor{red}{82.8\%} \\
		\bottomrule
	\end{tabularx}
	\caption{The performance of our proposed method on the 400 questions benchmark.}
	\label{tab:human_400}
\end{table*}

\begin{table*}[h!]
	\centering
	%\small 
	\footnotesize
	\begin{tabularx}{\textwidth}{l *{7}{>{\centering\arraybackslash}X}}
		\toprule
		\multicolumn{2}{c}{\bfseries Method} & 
		\multirow{2}{*}{\begin{tabular}[c]{@{}c@{}}\bfseries Logical\\ \bfseries Reasoning\end{tabular}} & 
		\multirow{2}{*}{\bfseries Accuracy} & 
		\multirow{2}{*}{\begin{tabular}[c]{@{}c@{}}\bfseries Comprehen\\ \bfseries -siveness\end{tabular}} & 
		\multirow{2}{*}{\bfseries Practicality} & 
		\multirow{2}{*}{\bfseries Score} & 
		\multirow{2}{*}{\begin{tabular}[c]{@{}c@{}} \bfseries Acceptable\\ \bfseries Rate\end{tabular}}\\  
		\cline{1-2}
		\bfseries X-Intelligence 3.0 &\bfseries RAG\\ \midrule
		
		\checkmark & $\times$  & 2.28 & 1.95 & 2.39 &2.22 &2.364 & 82.1\% \\
		
		\checkmark & \checkmark  & 2.4 & 2.07 & 2.57 &2.29 & \textcolor{red}{2.45} & \textcolor{red}{88.2\%} \\
		
		\bottomrule
	\end{tabularx}
	\caption{The performance of our domain-specific RAG on 400 questions benchmark.}
	\label{tab:human_195}
\end{table*}

\begin{table*}[h!]
	\centering
	\begin{tabularx}{\textwidth}{l *{2}{>{\centering\arraybackslash}X}}
		\toprule
		\textbf{800 Questions}  & \textbf{Deepseek-R1-671B} & \textbf{X-Intelligence 3.0} \\ \midrule
		
		Score & 2.373 & \textcolor{red}{2.431} \\
		
		Acceptable Rate & 81.3\% & \textcolor{red}{86.22\%}  \\
		\bottomrule
	\end{tabularx}
	\caption{The performance of our proposed method on the 800 questions benchmark.}
	\label{tab:human_800}
\end{table*}

To ensure a rigorous and professional evaluation process, we conducted assessments on sets of 100 and 400 questions with the participation of three domain experts holding doctoral degrees. For the 800-question evaluation, we broadened the scope by involving employees from various departments within the company, aiming to capture a wider range of real-world user perspectives.

All evaluations were performed using a blind review protocol with randomized presentation order. Identifying information was stripped from model outputs, which were then displayed in a random sequence to minimize evaluator bias. Evaluation criteria followed the guidelines detailed in Section \ref{Human_Evaluation}. To ensure consistency and comparability, all reviewers were provided with a standardized scoring rubric and illustrative examples. This multi-tiered evaluation framework effectively combines expert insights with end-user feedback, enhancing both the objectivity and representativeness of the results.

Initial expert assessments on the 100-question set revealed that both the R1-32B and Qwen3-32B models, following our post-training, outperformed the DeepSeek-R1-671B baseline, with notable gains in acceptability rates (Table \ref{tab:human_100}). The evaluation was then extended to 400 questions, focusing on the post-trained Qwen3-32B model. As shown in Table \ref{tab:human_400}, it continued to demonstrate superior performance over DeepSeek-R1-671B. 
Among the 400 questions, our proposed RAG method successfully retrieved relevant context slices for 195 of them, as shown in Table \ref{tab:human_195}. For these 195 questions, the use of our RAG approach further improved the quality of the model's responses.

Finally, the evaluation scale was expanded to 800 questions, with blind assessments conducted by employees across diverse company departments. As presented in Table \ref{tab:human_800}, our post-trained model consistently outperformed DeepSeek-R1-671B across this broader user group, validating its strong applicability and robustness in real-world deployment scenarios.

\subsection{Objective Evaluation}

To systematically measure the model's performance, we adopted the automated evaluation method proposed in Section \ref{Objective_Evaluation} to conduct objective assessments. This approach enables consistent and reproducible evaluations on large-scale datasets, avoiding subjective biases introduced by human factors. To verify the practical effectiveness of the objective evaluation method in two main application scenarios, we also conducted experimental validations separately for each scenario.

\subsubsection{Iterative Model Selection}

\begin{table*}[h!]
	\centering
	\begin{tabularx}{\textwidth}{l *{7}{>{\centering\arraybackslash}X}}
		\toprule
		\textbf{Model Name} & \textbf{Base Model} & \textbf{SFT Model 1} & \textbf{SFT Model 2}& \textbf{SFT Model 3}& \textbf{SFT Model 4}& \textbf{RL Model 1}&\textbf{RL Model 2} \\ \midrule
		
		Human Score & 2.308 &	2.332 &	2.393 &	2.376 &	2.353 &	2.386 &	2.435 \\
		
		Rank of Human Score & 7 &	6 &	2 &	4 &	5 &	3 &	1 \\
		
		Mean  Precision & 0.4564	 & 0.4714 &	0.4696 &	0.4747 &	0.4563 &	0.3533 &	0.399 \\
		
		Mean Recall & 0.3276 & 0.3465 & 0.3661 &	0.3453 &	0.3306 &	0.4076 &	0.4006\\
		
		Weighted  Precision and Recall & 0.3663 & 0.384 &	0.3971 &	0.3842 &	0.3683 &	0.3913 &	0.4001\\
		
		Obiective Evaluation Rank  & 7 &	5 &	2 & 4 & 6 &	3 &	1\\
		\bottomrule
	\end{tabularx}
	\caption{Objective evaluation results in the iterative model selection task.}
	\label{tab:objective evaluation results 1}
\end{table*}

In this scenario, the primary objective is to rapidly evaluate multiple model checkpoints generated during training, responses produced by models trained with different data configurations, and outputs resulting from various deployment parameters. This enables us to identify the model that delivers the best response quality and overall capability. Therefore, we selected the aforementioned test set of 100 questions for quick validation.

First, we utilized the DeepSeek-R1 in combination with manually written reference answers and relevant text slices retrieved from a domain-specific database to generate the ground-truth required by the evaluation system. Subsequently, we conducted a human evaluation to confirm that the average score of the generated ground-truth reached 2.78 out of 3, indicating near-perfect reference answers. Next, we refined the generated reference answers by removing quotations and summary sections based on semantic relevance, ensuring that only relevant and non-redundant content was retained for key point extraction.

We then employed the OpenAI-o4-mini to extract key points from the refined reference answers, which were stored for subsequent use. OpenAI-o4-mini was chosen as a balanced trade-off between evaluation speed and model performance. To maintain consistency, the same model was used in the later stage for determining whether the generated responses contained the expected key points. Finally, we invoked the OpenAI-o4-mini to evaluate the responses generated by different models using the prepared reference answers, extracted key points, and model outputs. The evaluation results are shown in Table \ref{tab:objective evaluation results 1}.

By comparing the overall ranking scores from human and automated evaluations, it can be observed that the two methods show a high degree of consistency. Discrepancies only occurred in the assessment of SFT model 1 and SFT model 4. Given that the automated evaluation method successfully identified the top three performing models even when the differences in model capabilities were subtle, we conclude that this approach is well-suited for the iterative model selection scenario and can be further validated on larger-scale test sets to assess its generalizability.

\subsubsection{Large-scale Evaluation}

\begin{table*}[h!]
	\centering
	\begin{tabularx}{\textwidth}{l *{4}{>{\centering\arraybackslash}X}}
		\toprule
		\textbf{Model Name} & \textbf{Model 1} & \textbf{DeepSeek-R1} & \textbf{Model 2}& \textbf{Model 3} \\ \midrule
		
		Human Score & 2.44&	2.316&	2.389&	2.447 \\
		
		Rank of Human Score & 2	&4&	3&	1 \\
		
		Rank by DeepSeek-R1&	2&	4&	3&	1 \\
		
		Rank by OpenAI-o4-mini&	3&	4&	2&	1\\
		
		Rank by OpenAI-o3&3&	4&	2&	1\\
		
		Rank by GPT-4.1	&4&	2&	3&	1\\
		\bottomrule
	\end{tabularx}
	\caption{Objective evaluation results of large-scale evaluation datasets which contain 100 questions.}
	\label{tab:objective evaluation results 2}
\end{table*}

\begin{table*}[h!]
	\centering
	\begin{tabularx}{\textwidth}{l *{3}{>{\centering\arraybackslash}X}}
		\toprule
		\textbf{Model Name} &  \textbf{DeepSeek-R1} & \textbf{Model 1} & \textbf{Model 2} \\  \midrule
		
		Human Score&	2.259&	2.403&	2.353 \\
		
		Rank of Human Score&	3&	1&	2\\
		
		Score of Objective Evaluation&	0.5226&	0.5818&	0.5399 \\
		
		Rank of Objective Evaluation &	3&	1&	2\\
		\bottomrule
	\end{tabularx}
	\caption{Objective evaluation results of large-scale evaluation datasets which contain 400 questions.}
	\label{tab:objective evaluation results 3}
\end{table*}

In this scenario, the goal of employing an automated evaluation method is to enhance efficiency and minimize reliance on human reviewers. 
Although fewer models are involved compared to the previous setting, the evaluation demands higher stability and reliability. 
To determine the most suitable LLM for this task, we compared the evaluation outcomes of four different LLMs across 100 test questions and four sets of model-generated responses.

The evaluation process closely followed the methodology used in the previous scenario. Reference answers were generated using DeepSeek-R1, while the remaining evaluation steps were carried out using the LLMs under comparison. The experimental results are presented in the Table \ref{tab:objective evaluation results 2}.

The results indicate that only when using DeepSeek-R1 for evaluation did the ranking of model capabilities align fully with human judgment. All other LLMs demonstrated varying degrees of deviation.
Notably, GPT-4.1 \cite{openai2025gpt-4.1} performed the weakest.
We hypothesize that this is due to the nature of the key point matching task, which requires strong reasoning ability. GPT-4.1 is not specifically optimized for reasoning tasks, which may explain its limitations in this context.
Furthermore, DeepSeek-R1 outperformed OpenAI models, likely because of its stronger comprehension and reasoning capabilities in Chinese-language tasks.

After identifying DeepSeek-R1 as the most suitable model for this task, we conducted a larger-scale evaluation using 400 test questions to assess three different models. In this round, human evaluation was performed through a blind review process, which contributed to greater accuracy and objectivity. The automated evaluation was carried out independently, without access to the human results, in order to reduce potential bias. The final comparison results are shown in Table \ref{tab:objective evaluation results 3}.

In summary, based on the above experimental evaluations, we conclude that the automated evaluation method introduced in this paper can effectively serve as a reliable alternative to human evaluation for specific tasks.

\subsection{RAG Human Evaluation}
To validate the performance improvements attributable to core retrieval modules (e.g., BGE embedding models, BGE rerank models), we employed a consistent Qwen-32B base model throughout the retrieval framework's progressive iterative optimization. 
Domain experts conducted manual evaluations, assessing the final performance across multiple dimensions through comparative experiments.

\begin{table*}[h!]
	\centering
	%\small 
	\footnotesize
	\begin{tabularx}{\textwidth}{l *{9}{>{\centering\arraybackslash}X}}
		\toprule
		\multicolumn{4}{c}{\bfseries method} & 
		\multirow{2}{*}{\begin{tabular}[c]{@{}c@{}}\bfseries logical\\ \bfseries reasoning\end{tabular}} & 
		\multirow{2}{*}{\bfseries accuracy} & 
		\multirow{2}{*}{\begin{tabular}[c]{@{}c@{}}\bfseries comprehen\\ \bfseries -siveness\end{tabular}} & 
		\multirow{2}{*}{\bfseries practicality} & 
		\multirow{2}{*}{\bfseries summary} & 
		\multirow{2}{*}{\begin{tabular}[c]{@{}c@{}}\bfseries acceptable\\ \bfseries rate\end{tabular}}\\  
		\cline{1-4}
		\bfseries baseline & \bfseries embedding Opt. & \bfseries rerank Opt. & \bfseries lterative search\\ \midrule
		
		\checkmark & $\times$ & $\times$ & $\times$ & 1.8821 & 1.647 & 1.824 & 1.750 & 1.994 & 50\%\\ 
		
		\checkmark & \checkmark & $\times$ & $\times$ & 2.353 & 2.000 & 2.118 & 2 & 2.259 & 64\% \\
		
		\checkmark & \checkmark & \checkmark & $\times$ & 2.294 & 1.941 & 2.353 &2.294 & 2.376 & 76\% \\
		
		\checkmark & \checkmark & \checkmark & \checkmark  & 2.156 & 1.956 & 2.556 & 2.467 & \textcolor{red}{2.458} & \textcolor{red}{82\%} \\
		
		\bottomrule
	\end{tabularx}
	\caption{Comparison of RAG core component optimization effects.}
	\label{tab:rag_1}
\end{table*}

The results of Table \ref{tab:rag_1} demonstrate that general-domain embedding and rerank models deliver suboptimal performance in vertical domains, indicating persistent knowledge misalignment when applying generic models to specialized contexts. After domain-specific fine-tuning, this knowledge misalignment is significantly mitigated, yielding measurable performance improvements. Furthermore, as evidenced in the last row of Table \ref{tab:rag_1}, iterative retrieval contributes additional gains. Analysis of questions showing substantial performance gains reveals that the algorithm's advantage stems primarily from enhanced retrieval effectiveness for complex multi-step reasoning problems, thus yielding overall score improvements.

\section{Conclusion and Future Research Directions}
\label{discussion_section}

This paper presents X-Intelligence 3.0, the first high-performance reasoning model specifically designed for the semiconductor display industry.
To support its development, we constructed a high-quality industry dataset and implemented a two-stage optimization process involving Supervised Fine-Tuning and Reinforcement Learning.
We also developed an automated evaluation framework that can effectively replace human experts in assessment, leading to significant improvements in both efficiency and consistency.
Furthermore, we integrated a domain-specific retrieval-augmented generation (RAG) method, which substantially enhances the model’s accuracy on evaluation benchmarks.
X-Intelligence 3.0 achieves exceptional performance across various domain-specific benchmarks, consistently surpassing DeepSeek-R1-67B. 
Future work will expand and refine the training corpus, incorporate additional pre-training, and deepen reinforcement-learning research to unlock even greater capabilities.

\newpage
\section*{Contributors and Acknowledgments}
\label{Contributors_Acknowledgments}

\subsection*{Contributors}
\noindent \textbf{TCL Corporate Research:} \\
Xiaolin Yan, Yangxing Liu\footnote{Corresponding author: yangxing.liu@tcl.com}, Jiazhang Zheng, Chi Liu, Mingyu Du, Caisheng Chen, Haoyang Liu, Ming Ding, Yuan Li, Qiuping Liao, Linfeng Li, Zhili Mei, Siyu Wan, Li Li, Ruyi Zhong, Jiangling Yu, Xule Liu, Huihui Hu, Jiameng Yue, Ruohui Cheng, Qi Yang, Liangqing Wu, Ke Zhu, Chi Zhang, Chufei Jing, Yifan Zhou, Yan Liang, Dongdong Li, Zhaohui Wang

\vspace{6pt}

\noindent \textbf{TCL China Star Optoelectronic Technology Co, Ltd.:} \\
Bin Zhao, Mingzhou Wu, Mingzhong Zhou, Peng Du, Zuomin Liao, Chao Dai, Pengfei Liang, Xiaoguang Zhu, Yu Zhang, Yu Gu, Kun Pan, Yuan Wu, Yanqing Guan, Shaojing Wu, Zikang Feng, Xianze Ma, Peishan Cheng, Wenjuan Jiang, Jing Ba, Huihao Yu, Zeping Hu, Yuan Xu, Zhiwei Liu, He Wang

\vspace{6pt}

\noindent \textbf{National Center of Technology Innovation for Display:} \\
Zhenguo Lin, Ming Liu, Yanhong Meng

\subsection*{Acknowledgments}
We would like to express our sincere gratitude to a key partner in the development of X-Intelligence 3.0—Alibaba Tongyi team. 
LLM experts from Tongyi team provided lots of valuable experience and useful guidance to help us address critical challenges we encountered during model research. 
Their exceptional expertise and collaborative spirit provided strong support for the smooth progress of this research.

\newpage
\bibliographystyle{unsrtnat}
\bibliography{references}

\begin{thebibliography}{32}
\providecommand{\natexlab}[1]{#1}
\providecommand{\url}[1]{\texttt{#1}}
\expandafter\ifx\csname urlstyle\endcsname\relax
  \providecommand{\doi}[1]{doi: #1}\else
  \providecommand{\doi}{doi: \begingroup \urlstyle{rm}\Url}\fi

\bibitem[Menozzi et~al.(2001)Menozzi, Lang, Naepflin, Zeller, and
  Krueger]{menozzi2001crt}
Marino Menozzi, Felicitas Lang, U~Naepflin, C~Zeller, and H~Krueger.
\newblock Crt versus lcd: Effects of refresh rate, display technology and
  background luminance in visual performance.
\newblock \emph{Displays}, 22\penalty0 (3):\penalty0 79--85, 2001.

\bibitem[Sobel(1991)]{sobel1991plasma}
Alan Sobel.
\newblock Plasma displays.
\newblock \emph{IEEE Transactions on plasma science}, 19\penalty0 (6):\penalty0
  1032--1047, 1991.

\bibitem[Schadt(1997)]{schadt1997liquid}
Martin Schadt.
\newblock Liquid crystal materials and liquid crystal displays.
\newblock \emph{Annual review of materials science}, 27\penalty0 (1):\penalty0
  305--379, 1997.

\bibitem[Song et~al.(2020)Song, Lee, Jeong, Choi, and Yoo]{song2020organic}
Jinouk Song, Hyeonwoo Lee, Eun~Gyo Jeong, Kyung~Cheol Choi, and Seunghyup Yoo.
\newblock Organic light-emitting diodes: pushing toward the limits and beyond.
\newblock \emph{Advanced Materials}, 32\penalty0 (35):\penalty0 1907539, 2020.

\bibitem[Dayneko et~al.(2016)Dayneko, Lypenko, Linkov, Sannikova, Samokhvalov,
  Nikitenko, and Chistyakov]{dayneko2016application}
Sergey Dayneko, Dmitriy Lypenko, Pavel Linkov, Nataliya Sannikova, Pavel
  Samokhvalov, Vladimir Nikitenko, and Alexander Chistyakov.
\newblock Application of cdse/zns/cds/zns core--multishell quantum dots to
  modern oled technology.
\newblock \emph{Materials Today: Proceedings}, 3\penalty0 (2):\penalty0
  211--215, 2016.

\bibitem[Huang et~al.(2019)Huang, Tan, Gou, Li, Lee, and
  Wu]{huang2019prospects}
Yuge Huang, Guanjun Tan, Fangwang Gou, Ming-Chun Li, Seok-Lyul Lee, and
  Shin-Tson Wu.
\newblock Prospects and challenges of mini-led and micro-led displays.
\newblock \emph{Journal of the Society for Information Display}, 27\penalty0
  (7):\penalty0 387--401, 2019.

\bibitem[Miao et~al.(2024)Miao, Hsiao, Sheng, Lee, Hong, Tsai, Chen, Liu, Lin,
  Chung, et~al.]{miao2024microdisplays}
Wen-Chien Miao, Fu-He Hsiao, Yujia Sheng, Tzu-Yi Lee, Yu-Heng Hong, Chun-Wei
  Tsai, Hung-Lung Chen, Zhaojun Liu, Chun-Liang Lin, Ren-Jei Chung, et~al.
\newblock Microdisplays: mini-led, micro-oled, and micro-led.
\newblock \emph{Advanced Optical Materials}, 12\penalty0 (7):\penalty0 2300112,
  2024.

\bibitem[Plaat et~al.(2024)Plaat, Wong, Verberne, Broekens, van Stein, and
  Back]{plaat2024reasoning}
Aske Plaat, Annie Wong, Suzan Verberne, Joost Broekens, Niki van Stein, and
  Thomas Back.
\newblock Reasoning with large language models, a survey.
\newblock \emph{arXiv preprint arXiv:2407.11511}, 2024.

\bibitem[Xu et~al.(2025)Xu, Hao, Zong, Wang, Zhang, Wang, Lan, Gong, Ouyang,
  Meng, et~al.]{xu2025towards}
Fengli Xu, Qianyue Hao, Zefang Zong, Jingwei Wang, Yunke Zhang, Jingyi Wang,
  Xiaochong Lan, Jiahui Gong, Tianjian Ouyang, Fanjin Meng, et~al.
\newblock Towards large reasoning models: A survey of reinforced reasoning with
  large language models.
\newblock \emph{arXiv preprint arXiv:2501.09686}, 2025.

\bibitem[{OpenAI}(2024)]{openai2024a}
{OpenAI}.
\newblock Learning to reason with llms, 2024.
\newblock URL \url{https://openai.com/index/learning-to-reason-with-llms/}.

\bibitem[Guo et~al.(2025)Guo, Yang, Zhang, Song, Zhang, Xu, Zhu, Ma, Wang, Bi,
  et~al.]{guo2025deepseek}
Daya Guo, Dejian Yang, Haowei Zhang, Junxiao Song, Ruoyu Zhang, Runxin Xu,
  Qihao Zhu, Shirong Ma, Peiyi Wang, Xiao Bi, et~al.
\newblock Deepseek-r1: Incentivizing reasoning capability in llms via
  reinforcement learning.
\newblock \emph{arXiv preprint arXiv:2501.12948}, 2025.

\bibitem[{Anthropic}(2025)]{anthropic2025claude37}
{Anthropic}.
\newblock Claude 3.7 sonnet: A hybrid reasoning model with dynamic thinking
  modes, 2025.
\newblock URL \url{https://www.anthropic.com/news/claude-3-7-sonnet}.

\bibitem[Yang et~al.(2025)Yang, Li, Yang, Zhang, Hui, Zheng, Yu, Gao, Huang,
  Lv, et~al.]{yang2025qwen3}
An~Yang, Anfeng Li, Baosong Yang, Beichen Zhang, Binyuan Hui, Bo~Zheng, Bowen
  Yu, Chang Gao, Chengen Huang, Chenxu Lv, et~al.
\newblock Qwen3 technical report.
\newblock \emph{arXiv preprint arXiv:2505.09388}, 2025.

\bibitem[{Google DeepMind}(2025)]{google2025gemini25}
{Google DeepMind}.
\newblock Gemini 2.5: Our most intelligent ai model, 2025.
\newblock URL
  \url{https://blog.google/technology/google-deepmind/gemini-model-thinking-updates-march-2025/}.

\bibitem[{LMSYS Org.}(2025)]{chatbot_arena_leaderboard}
{LMSYS Org.}
\newblock {Chatbot Arena Leaderboard}, 2025.
\newblock URL \url{https://lmarena.ai/leaderboard}.
\newblock Crowd-sourced Elo leaderboard for large language models.

\bibitem[Verma et~al.(2025)Verma, Zhou, Chandra, Kumar, and
  De~Choudhury]{verma2025framework}
Gaurav Verma, Jiawei Zhou, Mohit Chandra, Srijan Kumar, and Munmun
  De~Choudhury.
\newblock A framework for situating innovations, opportunities, and challenges
  in advancing vertical systems with large ai models.
\newblock \emph{arXiv preprint arXiv:2504.02793}, 2025.

\bibitem[{TCL}(2023)]{tcl2023x-intelligence1.0}
{TCL}.
\newblock Dtc2023: The first language model for semiconductor display sector,
  2023.
\newblock URL \url{https://mp.weixin.qq.com/s/KTEi-_HiTHni9kSRnDD5ig}.

\bibitem[{OpenAI}(2023)]{GPT4}
{OpenAI}.
\newblock Gpt-4 is openai’s most advanced system, producing safer and more
  useful responses, 2023.
\newblock URL \url{https://openai.com/index/gpt-4/}.

\bibitem[Wei et~al.(2022)Wei, Wang, Schuurmans, Bosma, Xia, Chi, Le, Zhou,
  et~al.]{wei2022chain}
Jason Wei, Xuezhi Wang, Dale Schuurmans, Maarten Bosma, Fei Xia, Ed~Chi, Quoc~V
  Le, Denny Zhou, et~al.
\newblock Chain-of-thought prompting elicits reasoning in large language
  models.
\newblock \emph{Advances in neural information processing systems},
  35:\penalty0 24824--24837, 2022.

\bibitem[Chen et~al.(2024)Chen, Xiao, Zhang, Luo, Lian, and Liu]{chen2024bge}
Jianlv Chen, Shitao Xiao, Peitian Zhang, Kun Luo, Defu Lian, and Zheng Liu.
\newblock Bge m3-embedding: Multi-lingual, multi-functionality,
  multi-granularity text embeddings through self-knowledge distillation.
\newblock \emph{arXiv preprint arXiv:2402.03216}, 2024.

\bibitem[Liu et~al.(2024)Liu, Feng, Xue, Wang, Wu, Lu, Zhao, Deng, Zhang, Ruan,
  et~al.]{liu2024deepseek}
Aixin Liu, Bei Feng, Bing Xue, Bingxuan Wang, Bochao Wu, Chengda Lu, Chenggang
  Zhao, Chengqi Deng, Chenyu Zhang, Chong Ruan, et~al.
\newblock Deepseek-v3 technical report.
\newblock \emph{arXiv preprint arXiv:2412.19437}, 2024.

\bibitem[{Hugging Face}(2025)]{huggingface_open_r1}
{Hugging Face}.
\newblock {Open r1: A Fully Open Reproduction of DeepSeek-R1}, January 2025.
\newblock URL \url{https://github.com/huggingface/open-r1}.
\newblock GitHub repository.

\bibitem[Tian et~al.(2025)Tian, Zhao, Wang, Chen, Peng, Ji, Zhao, and
  Li]{tian2025deepdistill}
Xiaoyu Tian, Sitong Zhao, Haotian Wang, Shuaiting Chen, Yiping Peng, Yunjie Ji,
  Han Zhao, and Xiangang Li.
\newblock Deepdistill: Enhancing llm reasoning capabilities via large-scale
  difficulty-graded data training.
\newblock \emph{arXiv preprint arXiv:2504.17565}, 2025.

\bibitem[Penedo et~al.(2025)Penedo, Lozhkov, Kydlíček, Ben~Allal, Beeching,
  Piqueres~Lajarín, Gallouédec, Habib, Tunstall, and von
  Werra]{penedo2025codeforcescots}
Guilherme Penedo, Anton Lozhkov, Hynek Kydlíček, Loubna Ben~Allal, Edward
  Beeching, Agustín Piqueres~Lajarín, Quentin Gallouédec, Nathan Habib,
  Lewis Tunstall, and Leandro von Werra.
\newblock Codeforces cots dataset, 2025.
\newblock URL \url{https://huggingface.co/datasets/open-r1/codeforces-cots}.
\newblock Hugging Face dataset.

\bibitem[Charikar(2002)]{charikar2002similarity}
Moses~S Charikar.
\newblock Similarity estimation techniques from rounding algorithms.
\newblock In \emph{Proceedings of the thiry-fourth annual ACM symposium on
  Theory of computing}, pages 380--388, 2002.

\bibitem[Rafailov et~al.(2023)Rafailov, Sharma, Mitchell, Manning, Ermon, and
  Finn]{rafailov2023direct}
Rafael Rafailov, Archit Sharma, Eric Mitchell, Christopher~D Manning, Stefano
  Ermon, and Chelsea Finn.
\newblock Direct preference optimization: Your language model is secretly a
  reward model.
\newblock \emph{Advances in Neural Information Processing Systems},
  36:\penalty0 53728--53741, 2023.

\bibitem[Huang and Huang(2024)]{huang2024survey}
Yizheng Huang and Jimmy Huang.
\newblock A survey on retrieval-augmented text generation for large language
  models.
\newblock \emph{arXiv preprint arXiv:2404.10981}, 2024.

\bibitem[Li et~al.(2024)Li, Tang, Chen, and Chen]{li2024conan}
Shiyu Li, Yang Tang, Shizhe Chen, and Xi~Chen.
\newblock Conan-embedding: General text embedding with more and better negative
  samples.
\newblock \emph{arXiv preprint arXiv:2408.15710}, 2024.

\bibitem[Qin et~al.(2025)Qin, Dong, Zhang, Dong, Huang, Yang, Khademi, Zhang,
  Awadalla, Fung, et~al.]{qin2025scaling}
Zeyu Qin, Qingxiu Dong, Xingxing Zhang, Li~Dong, Xiaolong Huang, Ziyi Yang,
  Mahmoud Khademi, Dongdong Zhang, Hany~Hassan Awadalla, Yi~R Fung, et~al.
\newblock Scaling laws of synthetic data for language models.
\newblock \emph{arXiv preprint arXiv:2503.19551}, 2025.

\bibitem[Es et~al.(2025)Es, James, Espinosa-Anke, and
  Schockaert]{es2025ragasautomatedevaluationretrieval}
Shahul Es, Jithin James, Luis Espinosa-Anke, and Steven Schockaert.
\newblock Ragas: Automated evaluation of retrieval augmented generation, 2025.
\newblock URL \url{https://arxiv.org/abs/2309.15217}.

\bibitem[{openai}(2025{\natexlab{a}})]{openai2025o4mini}
{openai}.
\newblock Introducing openai o3 and o4-mini, 2025{\natexlab{a}}.
\newblock URL \url{https://openai.com/index/introducing-o3-and-o4-mini/}.

\bibitem[{openai}(2025{\natexlab{b}})]{openai2025gpt-4.1}
{openai}.
\newblock Introducing gpt-4.1 in the api, 2025{\natexlab{b}}.
\newblock URL \url{https://openai.com/index/gpt-4-1/}.

\end{thebibliography}

\end{document}